\documentclass[runningheads]{llncs}
\usepackage[T1]{fontenc}

\usepackage{graphicx}
\usepackage{amsmath}
\usepackage{cite}
\usepackage{booktabs}
\usepackage{hyperref}
\usepackage{color}

\begin{document}
\title{Toward Improved Generalization: Meta Transfer of Self-supervised Knowledge on Graphs}
% \thanks{Supported by organization x.}
%
\titlerunning{Meta Transfer of Self-supervised Knowledge}
\author{Wenhui Cui, Haleh Akrami, Anand A. Joshi, Richard M. Leahy}   
% \and Second Author\inst{2,3}\orcidID{1111-2222-3333-4444} \and
\authorrunning{Cui et al.}
\institute{University of Southern California}
\maketitle              % typeset the header of the contribution
\begin{abstract}
Despite the remarkable success achieved by graph convolutional networks for functional brain activity analysis, the heterogeneity of functional patterns and the scarcity of imaging data still pose challenges in many tasks. Transferring knowledge from a source domain with abundant training data to a target domain is effective for improving representation learning on scarce training data. However, traditional transfer learning methods often fail to generalize the pre-trained knowledge to the target task due to domain discrepancy. Self-supervised learning on graphs can increase the generalizability of graph features since self-supervision concentrates on inherent graph properties that are not limited to a particular supervised task.
We propose a novel knowledge transfer strategy by integrating meta-learning with self-supervised learning to deal with the heterogeneity and scarcity of fMRI data. Specifically, we perform a self-supervised task on the source domain and apply meta-learning, which strongly improves the generalizability of the model using the bi-level optimization, to transfer the self-supervised knowledge to the target domain.  Through experiments on a neurological disorder classification task, we demonstrate that the proposed strategy significantly improves target task performance by increasing the generalizability and transferability of graph-based knowledge.

\end{abstract}
\section{Introduction}
Graph Convolutional Networks (GCNs) have shown strong performance in analyzing brain connectivity based on functional magnetic resonance imaging (fMRI) ~\cite{li2021braingnn, gadgil2020spatio,ahmedt2021graph}, but the scarcity and heterogeneity of fMRI data still pose challenges in domains like Attention-Deficit/Hyperactivity Disorder (ADHD)~\cite{huang2020multi} and post-traumatic epilepsy~\cite{akbar2022post}.
Transferring knowledge from a source domain with abundant training data to a target domain with a limited number of training samples can effectively improve representation learning on target-domain scarce training data~\cite{Han2021, chen2019catastrophic}. Approaches like fine-tuning~\cite{raghu2019transfusion}, joint training~\cite{ge2017borrowing}, and multi-task learning~\cite{kendall2018multi} are used to utilize knowledge across different domains. However, traditional transfer learning methods often fail to generalize on target tasks due to domain discrepancy~\cite{liu2020metalearning, raghu2019transfusion}.  A desirable training strategy should learn domain-general features from the source domain and enable beneficial knowledge transfer to the target domain. In the widely applied fine-tuning scheme, the model is mostly pre-trained in a supervised manner that encodes source-domain-specific features. These features might be irrelevant to the target domain and this process may distort the learning of target-domain-specific features~\cite{chen2019catastrophic}. Self-supervised tasks have shown the ability to enhance the generalization of features in the presence of domain shifts~\cite{azizi2021big,reed2022self}. In contrast to supervised problems such as classification or segmentation, self-supervised frameworks are usually designed to learn intrinsic features like graph properties or image context that are not restricted to a specific supervised task~\cite{taleb20203d, wei2022masked}. Here, we apply a self-supervised representation learning method on the source domain to learn more generalizable features.

The heterogeneity of fMRI data is another factor affecting the generalization of graph-based knowledge.
Meta-learning has gained tremendous attention recently because of its learning-to-learn mechanism, which strongly increases the generalizability of models across different tasks~\cite{zhang2019metapred, liu2020meta, finn2017model, liu2021metacon}. The Model Agnostic Meta-Learning method (MAML)~\cite{finn2017model} is a gradient-based approach that uses a bi-level optimization scheme to enable the model to learn how to generalize on an unseen domain during training and has achieved remarkable success in few-shot learning tasks. Instead of learning to generalize from multiple source tasks to multiple target tasks in MAML,~\cite{liu2020meta} proposed a meta representation learning approach to learn transferable features from one source domain and improve the knowledge generalization to a target domain. 
% In this work, we investigate knowledge transfer strategies only in scenarios where we have access to source domain data.

We propose Meta Transfer of Self-supervised Knowledge (MeTSK), a unique meta-learning-based mechanism to transfer self-supervised knowledge on graphs. Meta-learning and self-supervised learning are two techniques improving the generalization of learned representations. Both the target task and the self-supervised source task are trained concurrently. 
Our network architecture consists of a feature extractor that identifies general features from both domains and source and target heads to learn domain-specific features for the source domain and target domain, respectively. The bi-level optimization is applied to learn transferable and generalizable graph features using a Spatio-temporal Graph Convolutional Network (ST-GCN)~\cite{gadgil2020spatio} as the backbone model.
Our experimental results demonstrate the effectiveness of the proposed strategy on a neurological disorder (ADHD) classification task.

In summary, our contribution is two-fold: (i) addressing data scarcity by introducing knowledge transfer from a source domain with abundant data to the target domain;
(ii) addressing data heterogeneity through integrating meta-learning with a self-supervised source task that jointly improve the generalization of graph-based knowledge under domain shifts.

\section{Methods}

\begin{figure}
    \centering
\includegraphics[width=0.95\textwidth]{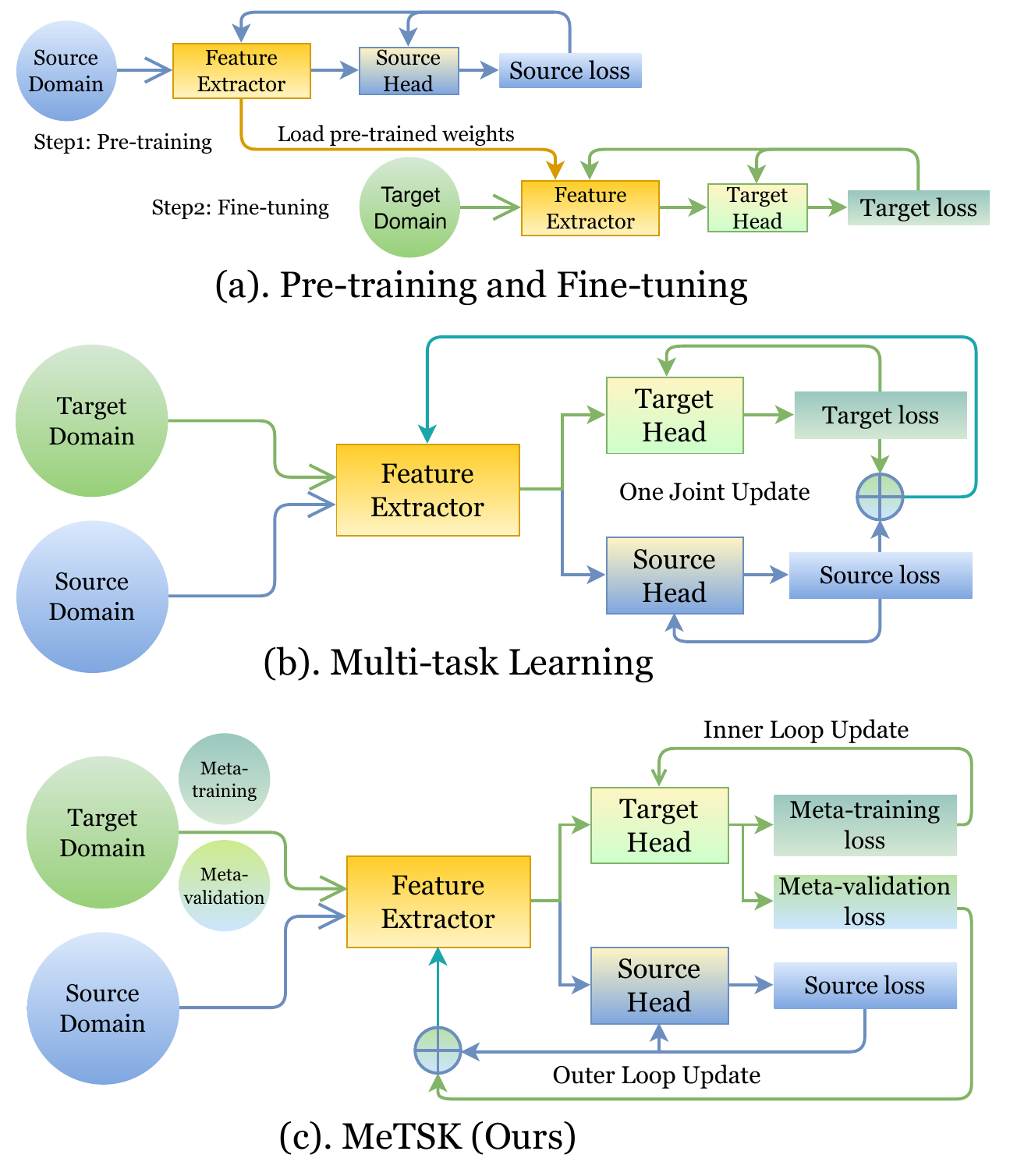}
    \caption{An illustration of the proposed meta-knowledge transfer strategy along with fine-tuning and multi-task learning methods: (a) shows the classic fine-tuning method: first pre-train the model on the source domain and then use the same pre-trained feature extractor for target domain training, with the head re-initialized; (b) shows the multi-task learning strategy to train on source and target domain simultaneously and only backpropagate once using gradients calculated from the total loss; (c) shows the proposed strategy MeTSK, where two optimization loops are involved in training. The inner loop only updates the target head, while the outer loop updates the source head and feature extractor. }
    \label{fig1}
\end{figure}

We first introduce the proposed meta-learning-based strategy, MeTSK, which improves the effectiveness of knowledge transfer on graphs from a self-supervised source task to a target task and increases the generalization of graph features. Assume there exists a source domain $\mathcal{S}$ with abundant training data $X_\mathcal{S}$ and a target domain $\mathcal{T}$, where the training data $X_\mathcal{T}$ is limited. A feature extractor $f(\phi)$, a target head $h_\mathcal{T}(\theta_t)$, and a source head $h_\mathcal{S}(\theta_s)$ are constructed to learn features and generate source predictions $h_\mathcal{S}(f(X_\mathcal{S}; \phi); \theta_s)$ as well as target predictions $h_\mathcal{T}(f(X_\mathcal{T}; \phi); \theta_t)$, where $\phi$, $\theta_t$, and $\theta_s$ are model parameters.
%The optimization objective of source task is $\mathcal{L_S}$ and the optimization objective of target task is $\mathcal{L_T}$. 
We adopt ST-GCN~\cite{gadgil2020spatio} as the backbone architecture to extract graph representations from both spatial information and temporal information. A graph convolution and a temporal convolution are performed in one ST-GCN module. The feature extractor includes three ST-GCN modules. The target head and the source head share the same architecture, which consists of one ST-GCN module and one fully-connected layer.
The overall framework of MeTSK is illustrated in Fig.~\ref{fig1}(c).

\subsection{Spatio-temporal Graph Convolution}
To construct the graph, we treat brain regions parcellated by a brain atlas \cite{glasser2016multi} as the nodes and define edges using the functional connectivity between pairs of nodes measured by Pearson's correlation coefficient~\cite{bellec2017neuro}. We randomly sample sub-sequences from the whole fMRI time series to increase the size of training data by constructing multiple input graphs containing dynamic temporal information. The input graph for the $r$-th sub-sequence sample from the $n$-th subject $X^{(n,r)}$ has dimension $P\times L\times C_0$, where $P$ is the number of brain regions or parcels (nodes), $L$ is the length of the sampled sub-sequence, and $C_0$ is the length of input node feature.
In ST-GCN, a graph convolution~\cite{kipf2016semi}, applied to the spatial graph at time point $l$~\cite{gadgil2020spatio}, can be expressed as follows.
\begin{equation}
    \Tilde{X}^{(n, r, l)} = D^{-1/2}(A+I)D^{-1/2}X^{(n, r, l)}W_{C_0 \times C_1}
\end{equation}
where $A$ is the adjacency matrix consisting of edge weights, $I$ is the identity matrix, $D$ is a diagonal matrix such that $D_{ii}=\sum_j A_{ij} + 1$, $W$ is the trainable weight matrix, $L$ is the input node feature length and $C_1$ is output node feature length.
We apply temporal convolution to $\Tilde{X}^{(n, r, v)}$ over length-$C_1$ node features on each node $v$.

\subsection{Meta Knowledge Transfer}
Following the bi-level optimization strategy in~\cite{finn2017model, liu2020meta}, the gradient-based update of model parameters is performed in two optimization loops. The model first back propagates the gradients through the target head only in several fast adaptation steps, and then back propagates through the source head and feature extractor. The algorithm can be summarized in three steps as follows: 
\begin{enumerate}
    \item \textbf{Outer loop} ($M$ iterations): first initialize the target head $h_\mathcal{T}(\theta_t)$, then randomly sample target meta-training set $X_{\mathcal{T}_{tr}}$ and meta-validation set $X_{\mathcal{T}_{val}}$ from $X_\mathcal{T}$, where $X_{\mathcal{T}_{tr}} \bigcap X_{\mathcal{T}_{val}} = \emptyset$, $X_{\mathcal{T}_{tr}} \bigcup X_{\mathcal{T}_{val}} = X_\mathcal{T}$. One forward propagation is performed on the feature extractor to generate the graph representation $f(X_{\mathcal{T}_{tr}}; \phi)$.
    
    \item \textbf{Inner loop} ($k$ update steps): Only the target head parameters $\theta_t$ are updated using gradients computed from the optimization objective $\mathcal{L_T}$ of the target task. 
    \begin{equation}
        \theta_t^{j+1} = \theta_t^{j} - \alpha \nabla_{\theta_t^{j}} \mathcal{L_T} (h_\mathcal{T}(f(X_{\mathcal{T}_{tr}}; \phi^i); \theta_t^j))
    \end{equation}
    where $\alpha$ is the inner loop learning rate, and $\theta_t^j$ is the target head parameter at $j$-th update step.
    
    \item \textbf{Outer loop}: After the inner loop is finished, generate $h_\mathcal{T}(f(X_{\mathcal{T}_{val}}; \phi); \theta_t)$ using forward propagation through the feature extractor and target head to evaluate the generalizability of the target head on the validation set. Train the source task on $X_\mathcal{S}$ using the optimization objective $\mathcal{L_S}$. Freeze the target head in the outer loop and only update feature extractor parameters $\phi$ and source head parameters $\theta_s$:

    \begin{equation}
    \begin{split}
    \{\theta_s^{i+1}, \phi^{i+1}\} =
         \{\theta_s^{i}, \phi^{i} \} - \beta (\nabla_{\theta_s^{i}, \phi^i} \mathcal{L_S} (h_\mathcal{S}(f(X_{\mathcal{S}}; \phi^i); \theta_s^i)) \\
         +  \nabla_{\phi^i} \lambda \mathcal{L_T} (h_\mathcal{T}(f(X_{\mathcal{T}_{val}}; \phi^i); \theta_t^k)))
    \end{split}
    \end{equation}
    where $\beta$ is the outer loop learning rate, and $\lambda$ is a scaling coefficient.
 
\end{enumerate}
In the inner loop, the model first learns target-specific features in the target head. With $\mathcal{L_T}$ and $\mathcal{L_S}$ computed on $X_{\mathcal{T}_{val}}$ and $X_\mathcal{S}$ respectively in the outer loop, minimizing the total loss $\mathcal{L_S} + \lambda \mathcal{L_T}$ forces the model to learn transferable and generalizable features from the source domain.
\subsection{Self-supervised Learning}
Using a self-supervised task to distill knowledge from the source domain has proven to be effective for improving generalization to the target domain~\cite{wei2022masked,reed2022self, taleb20203d}. To further boost the generalizability of graph features we use a meta-learning framework and apply a graph contrastive loss~\cite{you2020graph} to perform a self-supervised task on the source domain. To construct input graphs for ST-GCN, we randomly sample sub-sequences $X^{(n,r_1)}$, $X^{(n,r_2)}$ $(r1 \neq r2)$ from the whole fMRI time series for subject $n$, which can be viewed as a form of data augmentation. $X^{(n,r_1)}$ and $X^{(n,r_2)}$ should produce similar graph features even though they contain different temporal information. The graph contrastive loss enforces similarity between graph features extracted from the same subject and dissimilarity between graph features extracted from different subjects~\cite{chen2020simple}. A cosine similarity is applied to measure the similarity in the latent graph feature space~\cite{you2020graph}.

\begin{equation}
        \mathcal{L_S} = \dfrac{1}{N} \sum_{n=1}^{N} -\log\frac{{\exp{( sim(\Tilde{X}_\mathcal{S}, n, n)/ \tau)}}}{\sum_{m=1, m\neq n}^{N} \exp{(sim(\Tilde{X}_\mathcal{S}, n, m)/ \tau)}}
\end{equation}
\begin{equation}
    sim(\Tilde{X}, n, m) = \frac{(\Tilde{X}^{(n, r_1)})^\top \Tilde{X}^{(m, r_2)}}  { \|\Tilde{X}^{(n, r_1)}\| \|\Tilde{X}^{(m, r_2)}\|}
\end{equation}
where $\Tilde{X}_\mathcal{S} = h_\mathcal{S}(f(X_\mathcal{S}; \phi); \theta_s)$ is the generated graph representation, $\tau$ is a temperature hyper-parameter, and N is the total number of subjects in one training batch. By minimizing the graph contrastive loss on the source domain, the model learns intrinsic graph features that provide generalizable knowledge across domains.

The optimization objective $\mathcal{L_T}$ of the target domain depends on the target task. In a classification task with class labels $Y_\mathcal{T}$, we adopt the Cross-Entropy loss. The total loss for the proposed meta-learning-based strategy, MeTSK, is
\begin{equation}
    \mathcal{L}_{meta} = \mathcal{L_S} - \lambda \sum_{\mathrm{classes}} Y_{\mathcal{T}}log(h_\mathcal{T}(f(X_{\mathcal{T}}; \phi); \theta_t) )
\end{equation}

\section{Experiments and Results}
\subsection{Datasets}
To evaluate the effectiveness of our proposed strategy, MeTSK, on fMRI data, we use the Human Connectome Project (HCP) S1200 dataset~\cite{van2013wu} as the source domain and the Attention-Deficit/Hyperactivity Disorder (ADHD-200) consortium data at Peking site~\cite{bellec2017neuro} as the target domain.
HCP includes 1,096 young adults (ages 22-35) subjects with resting-state-fMRI data with a total of 1200 time-points per session. The preprocessing of fMRI follows the minimal preprocessing procedure in~\cite{gadgil2020spatio, glasser2013minimal}. We then applied BrainSync~\cite{joshi2017brainsync,akrami2019group} to temporally align the fMRI time series across all subjects. Brainsync exploits similarity in the fMRI's spatial correlation structure across subjects to define an orthogonal (rotation and possible reflection) transformation matrix. As a result of this transformation, the time series are approximately aligned (highly-correlated) across subjects at homologous locations. Finally, the cortical surface of the brain was parcellated into $22$ ROIs using the surface alas in~\cite{glasser2016multi}. These 22 regions form the nodes of our graph.  The fMRI data were reduced to a single time-series per node by averaging across each ROI.

The target domain dataset ADHD-200 at Peking site includes 245 subjects in total, with 102 ADHD subjects and 143 Typically Developed Controls (TDC). 
% The 245 resting-state-fMRI data with 235 time points are pre-processed through normalization, and brainsync~\cite{joshi2018you}.
The rs-fMRI data was processed using BrainSuite (\url{https://brainsuite.org}) and its Functional Pipeline (BFP). We mapped the fMRI data onto the same grayordinate representation used for the HCP data\cite{SMITH2013666, BARCH2013169}. Preprocessing also largely followed the minimal-proprocessing scheme used for the HCP data, but as implemented in BrainSuite's BFP. The pre-processed fMRI data had $235$ time points that were aligned using Brainsync. As a final step, the data were transferred to the same 22-ROI cortical-surface atlas as for the HCP subjects, and the average time-series was computed for each ROI.

% The BrainSuite software as well as its functional processing and statistical analysis pipelines are publically available online. BFP is a software workflow that processes fMRI and T1 data using a combination of software that includes BrainSuite (\url{https://brainsuite.org}), AFNI (\url{https://afni.nimh.nih.gov}), FSL (\url{https://www.fmrib.ox.ac.uk/fsl}), and MATLAB scripts. Unique features of the BFP pipeline include cortically-constrained volumetric registration \cite{joshi2007surface, joshi_method2012}, Global PDF-based non-local means filtering (GPDF) \cite{Li_2020_MedImageAnal_Temporal,gpdf_jianli2018} and BrainSync alignment of resting fMRI time series \cite{JOSHI2018740,joshi2017brainsync}. 

% Starting from raw T1 images, BFP uses BrainSuite to perform brain extraction, tissue classification, generation of brain surfaces and coregistration to a reference anatomical atlas. fMRI processing includes motion correction, skull stripping, grand mean scaling, temporal filtering, detrending, spatial smoothing, nuisance signal regression and GPDF filtering. fMRI images are coregistered to T1 images and then transformed onto atlas space.

% BFP produces processed fMRI data represented both on surface and volume co-registered with BrainSuite's BCI-DNI atlas \cite{joshi2022hybrid} as well as a grayordinate based representation . 

% Secondly, the volumetric space generated by BFP comprise of a $51\times 70\times 70$ voxel-matrix volume at 3mm isotropic resolution in BCI-DNI atlas space \cite{joshi2022hybrid}.

\subsection{Results}
For comprehensive analysis and comparison, we designed (i) a baseline model with a supervised task directly trained on the ADHD data (Baseline),(ii) a model fine-tuned on ADHD data after pre-training on HCP data (FT) using the graph contrastive loss, (iii) a model performing multi-task learning on HCP data and ADHD data simultaneously (MTL), and (iv) the proposed meta knowledge transfer strategy, MeTSK. For MTL, we simply remove the inner loop in MeTSK and use all the training data to update the target head in the outer loop. As a result, both heads and the feature extractor are updated simultaneously in one loop. The features learned in the feature extractor are shared across two domains, and each head still encodes task-specific information.

\begin{figure}
    \centering
\includegraphics[width=0.85\textwidth]{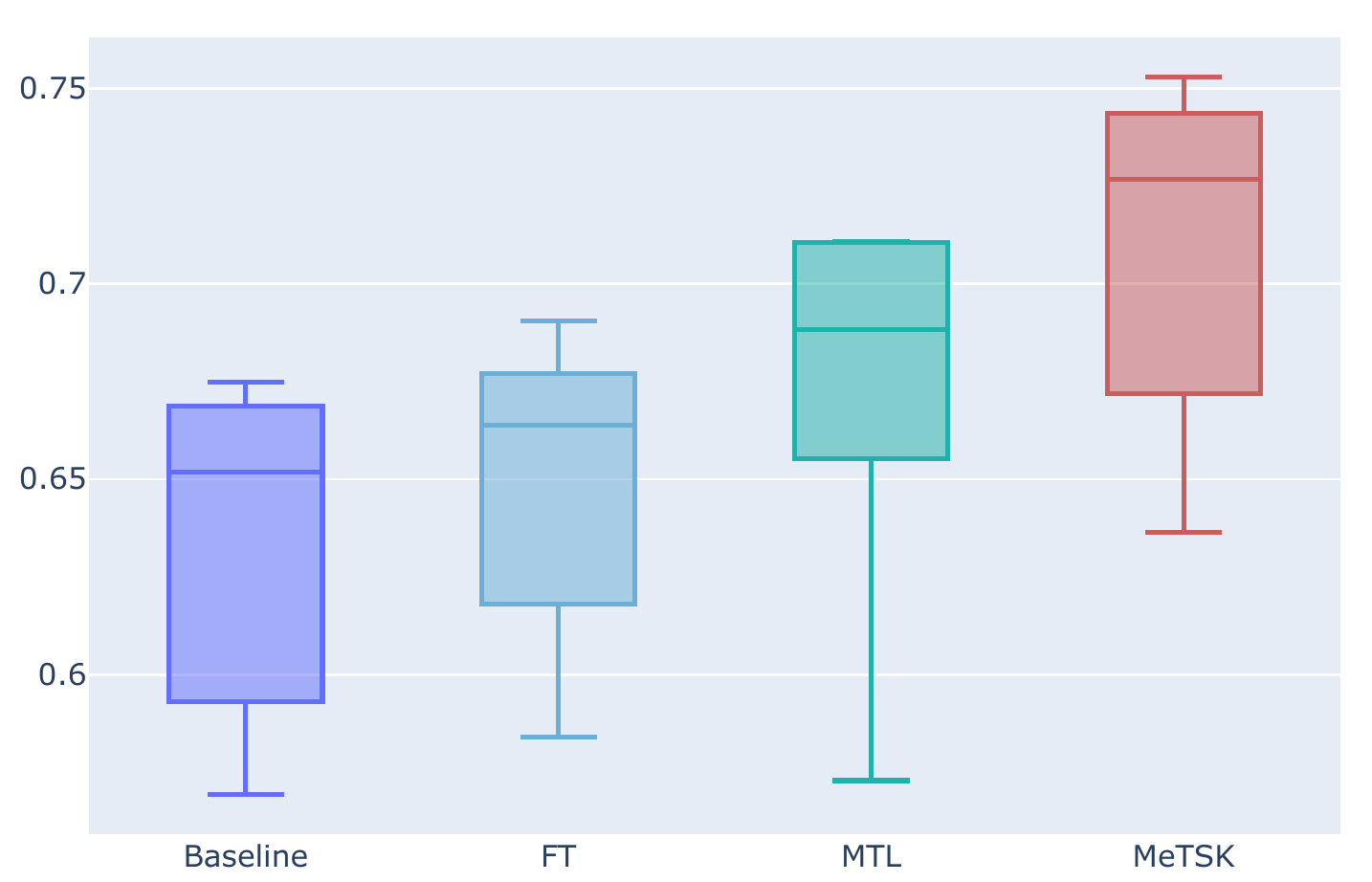}
    \caption{AUCs of 5-fold cross-validation on the testing set of ADHD data: a comparison of baseline, fine-tuning, multi-task learning, and the proposed strategy MeTSK with a self-supervised source task.}
    \label{fig:boxplot}
\end{figure}

We use 5-fold cross-validation to split training/testing sets on ADHD data and use all HCP data for training. In order to perform bi-level optimization, the training set of ADHD data in each fold is further divided into a meta-training set $X_{\mathcal{T}_{tr}}$ of $157$ subjects and a meta-validation set $X_{\mathcal{T}_{val}}$ of $39$ subjects. Model performance is evaluated on the test ADHD data set using the average area-under-the-ROC-curve (AUC) as the evaluation metric as shown in In Table~\ref{res} and in Fig~\ref{fig:boxplot}. The proposed method, MeTSK, achieved the best mean AUC of $0.7081$, which is a significant improvement compared to the baseline model which uses only ADHD data for training. This result demonstrated the effectiveness of MeTSK for handling data scarcity. MeTSK also surpassed the performance of traditional knowledge transfer approaches using fine-tuning and multi-task learning, providing evidence for the improved generalization and the advanced transfer of graph features. %The large standard deviation for the AUC in all cases likely reflects the heterogeneity of ADHD data. %suggest we delete this commented sentence. 

Performance improvement by fine-tuning a pre-trained model is only marginal because ADHD data is small-scale compared to HCP data and it is difficult for the pre-trained model to adapt to a new domain where knowledge is limited. Multi-task learning showed better performance due to feature sharing between the source domain and target domain during training. In this way the model can learn features that are beneficial for both domains instead of overfitting to one.
\begin{table}
\caption{A comparison of mean AUCs of 5-fold cross-validation on the testing set of ADHD data using different methods. A baseline model and a meta-learning model (MeL) were trained only on target data (discussed in section 3.4), which are shown in the left columns. The FT, MTL, and MeTSK methods are compared for two cases - transferring knowledge from (i) a self-supervised source task and (ii) a sex classification source task (discussed in section 3.4), respectively.}\label{res}
\centering
\begin{tabular}{c c||c c c}
\toprule[1pt]
\hline
 &  Target-only & Source \& Target & Self-supervision & Sex Classification\\
\hline
Baseline & $0.6327 \pm 0.0495$  & FT
 & $0.6487 \pm 0.0435$ & $0.6281 \pm 0.0539$ \\
MeL & $0.6585\pm  0.0297$ & MTL  & $0.6731 \pm 0.0513 $ & $0.6375 \pm 0.0572$\\

& &\textbf{MeTSK (ours)}& $\mathbf{0.7081 \pm 0.0427}$ & $0.6825 \pm 0.0457$\\
\hline
\bottomrule
\end{tabular}
\end{table}

\subsection{Generalizability of Learned Features} %Analysis}
\begin{figure}
    \centering
    \includegraphics[width=0.6\textwidth]{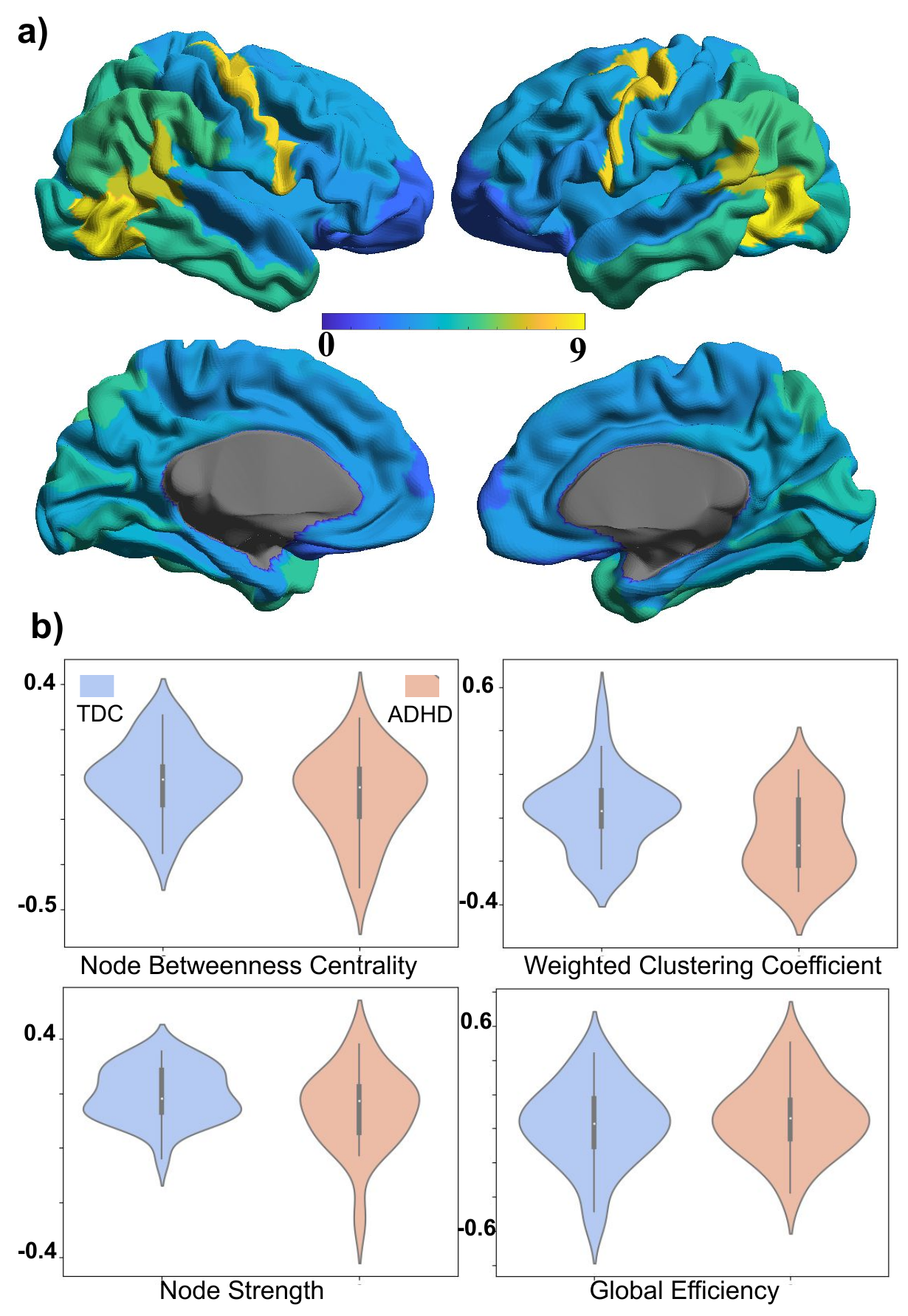}
    \caption{a) Mapping node importance to 22 ROIs of the cortical surface on a brain b) Pearson correlation of four node properties of the input graph (node betweenness centrality, the weighted clustering coefficient, node strength, and global efficiency) with node importance learned by the feature extractor.}
    \label{fig:graph}
\end{figure}

%feature visualization on training and testing
%Domain similarity analysis probably.
%Graph theory and properties, figure
The ST-GCN model learns an edge-importance matrix applied to graph convolutions as an attention mechanism during training~\cite{gadgil2020spatio}. We summed the importance weights of the edges connected to each node to compute the node importance and then mapped these node importance measures to the corresponding cortical regions in the brain atlas in Fig~\ref{fig:graph} (a). MT+\_Complex\_and\_Neighboring \_Visual\_Areas, 
 Premotor\_Cortex, and Temporal-Parietal-Occipital\_Junction are the most important cortex regions discovered by the last layer in feature extractor.

For further analysis, we calculated four nodal properties of the input graph that are typically meaningful in a fully connected weighted graph: node betweenness centrality, the weighted clustering coefficient, node strength, and global efficiency (all calculated using the Brain Connectivity Toolbox, \url{https://github.com/aestrivex/bctpy}). We then calculated the Pearson correlation of each of these features with the node importance for the test dataset to examine whether node importance captures known intrinsic and generalizable node features. Pearson correlation for the TDC and ADHD groups are shown in Fig \ref{fig:graph} b. There is a significant correlation between node importance and node strength for both groups (0.215), indicating that the learned features are, at least to some degree, generalizable. For further evaluation of the generalizability of the features generated by the feature extractor, we calculated the canonical correlation between the learned graph features and the four hand-crafted features. Canonical correlation analysis (CCA) is a method for exploring the relationships between two multivariate sets of variables (vectors), all measured on the same individual\cite{thompson2000canonical}. We reduced the dimension of the extracted graph features using PCA (principal component analysis) before applying CCA to ensure the number of features is less than the number of (subjects$\times$nodes). The maximum canonical-correlation value was 0.86, again indicating some consistency between general graph properties and automatically extracted features. 
%\begin{figure}
  %  \centering
%\includegraphics[width=0.75\textwidth]{ROI_importance.png}
  %  \caption{Mapping node importance to 22 ROIs of the cortical surface on a brain}
  %  \label{fig:node}
%\end{figure}

\subsection{Ablation Study}
We examine the individual contributions of self-supervised learning and meta-learning to generalizable knowledge transfer in this section. 
To explore the effect of meta-learning, we designed an experiment where only the target dataset was used in meta-learning, which means that the source head and the source loss are removed during the bi-level optimization. The target head is first trained to learn features from the meta-training set of ADHD data in the inner loop. Then the feature extractor learns how to produce features that generalize well on a held-out validation set in the outer loop. The performance of the meta-learning model (MeL) compared to the baseline model trained on target-only data is shown in Table~\ref{res}. The mean AUC improved from $0.6327$ to $0.6585$ even without source domain knowledge, which demonstrated the increased generalization achieved by meta-learning on the heterogeneous ADHD data.

To explore the contribution of self-supervised learning, we compared the effect of using a self-supervised task with using a sex classification task on the HCP dataset.
We implemented fine-tuning, multi-task learning, and MeTSK using sex classification (female vs male) as the source task. The same 5-fold cross-validation is applied to compare the average AUC. In Table~\ref{res}, FT, MTL, and MeTSK all showed a degraded performance when transferring knowledge from the sex classification task, where the learned knowledge is more task-specific and less generalizable. FT even achieved worse AUC than the baseline, which indicates that the sex-related features of the brain could be irrelevant to ADHD classification, thus harming the fine-tuned model performance.

\subsection{Implementation Details}
To optimize  model performance, all the methods including Baseline, FT, MTL, and MeTSK follow the training setting in~\cite{gadgil2020spatio} for the ST-GCN model. For meta-learning training, the inner loop update step is $30$, while the number of iterations in the outer loop is $3600$. We generate one meta-training batch by randomly selecting an equal number of samples from each class. The batch size is 32, both for the meta-training and the meta-validation set. We use an Adam optimizer~\cite{kingma2014adam} with learning rate $\beta=0.001$ in the outer loop, and an SGD optimizer~\cite{ketkar2017stochastic} with learning rate $\alpha=0.01$ in the inner loop. 
We set the hyper-parameter $\lambda=15$ and the temperature parameter $\tau=30$ to adjust the scale of losses following~\cite{liu2020meta, you2020graph}. 
Since contrastive loss converges slowly\cite{jaiswal2020survey}, a warm-up phase is applied to first train the model only on HCP data using the graph contrastive loss for $1800$ iterations. 
% The standard deviation of running cross-validation experiment for multiple times is around $0.02$.

% \begin{figure}
% \includegraphics[width=\textwidth]{fig1.eps}
% \caption{A figure caption is always placed below the illustration.
% Please note that short captions are centered, while long ones are
% justified by the macro package automatically.} \label{fig1}
% \end{figure}

%
% the environments 'definition', 'lemma', 'proposition', 'corollary',
% 'remark', and 'example' are defined in the LLNCS documentclass as well.
%
\section{Conclusion}
To tackle the heterogeneity and scarcity of fMRI data,
we propose a novel knowledge transfer strategy by integrating meta-learning with self-supervised learning. Specifically, we perform a self-supervised task on the source domain and apply meta-learning to transfer this learned knowledge to the target domain. Through experiments on the ADHD classification task, we demonstrated that the proposed strategy significantly improved target task performance by increasing the generalizability and the transferability of the graph-based knowledge.


\bibliographystyle{splncs04}
\bibliography{mybib}

\end{document}